\documentclass[runningheads]{llncs}
\usepackage[T1]{fontenc}
\usepackage{graphicx}
\usepackage{cite}
\usepackage{color,array}
\usepackage{amsfonts}
\usepackage{amsmath}
\usepackage{graphicx}
\usepackage{multirow}
\usepackage{adjustbox}
\usepackage[normalem]{ulem}
\useunder{\uline}{\ul}{}
\usepackage[hidelinks,colorlinks=true,linkcolor=red,citecolor=blue]{hyperref}
\usepackage{orcidlink}

\begin{document}
\title{FaR: Enhancing Multi-Concept Text-to-Image Diffusion via Concept Fusion and Localized Refinement}

\author{Gia-Nghia Tran\inst{1,4}\orcidlink{0009-0002-1889-1465} \and Quang-Huy Che\inst{1,4}\orcidlink{0009-0007-7477-4702} \and
Trong-Tai Dam Vu\inst{1,4}\orcidlink{0009-0005-1549-989X} \and
Bich-Nga Pham\inst{1,4}\orcidlink{0009-0008-3031-1711} \and
Vinh-Tiep Nguyen\inst{1,4}\orcidlink{0000-0003-4260-7874} \and
Trung-Nghia Le\inst{2,4}\orcidlink{0000-0002-7363-2610} \and
Minh-Triet Tran\inst{2,3,4}\orcidlink{0000-0003-3046-3041}}

\authorrunning{Gia-Nghia Tran et al.}

\institute{University of Information Technology, Ho Chi Minh City, Vietnam \email{\{nghiatg,huycq,taidvt,ngaptb,tiepnv\}@uit.edu.vn} \and
University of Science, Ho Chi Minh City, Vietnam \email{\{ltnghia,tmtriet\}@fit.hcmus.edu.vn} \and John von Neumann Institute, Ho Chi Minh City, Vietnam \and Vietnam National University, Ho Chi Minh City, Vietnam}

\titlerunning{Fuse-and-Refine}

\maketitle              
\begin{abstract}

Generating multiple new concepts remains a challenging problem in the text-to-image task. 
Current methods often overfit when trained on a small number of samples and struggle with attribute leakage, particularly for class-similar subjects (e.g., two specific dogs).
In this paper, we introduce Fuse-and-Refine (FaR), a novel approach that tackles these challenges through two key contributions: Concept Fusion technique and Localized Refinement loss function.
Concept Fusion systematically augments the training data by separating reference subjects from backgrounds and recombining them into composite images to increase diversity.
This augmentation technique tackles the overfitting problem by mitigating the narrow distribution of the limited training samples. 
In addition, Localized Refinement loss function is introduced to preserve subject representative attributes by aligning each concept’s attention map to its correct region.
This approach effectively prevents attribute leakage by ensuring that the diffusion model distinguishes similar subjects without mixing their attention maps during the denoising process.
By fine-tuning specific modules at the same time, FaR balances the learning of new concepts with the retention of previously learned knowledge.
Empirical results show that FaR not only prevents overfitting and attribute leakage while maintaining photorealism, but also outperforms other state-of-the-art methods.

\end{abstract}

\section{Introduction}
Text-to-image diffusion models \cite{diffusion, stablediffusion, fastdiffuser} have demonstrated significant advancements in producing high-resolution and realistic images. 
Based on these models, personalization techniques have also evolved. Several methods \cite{dreambooth, textualinversion, customdiffusion, svdiff, mixofshow, loracomposer} allow the models to generate images of a user-defined subject in novel contexts using just a few samples. 
Although current customization methods have achieved significant progress in single-concept scenarios, they face challenges when involving multiple concepts \cite{customdiffusion,svdiff,mixofshow,loracomposer}. These methods often suffer from two critical challenges: overfitting and attribute leakage.

The overfitting problem in diffusion models arises from the limited training data for each subject, which reduces the diversity of generated outputs. 
Since each training image usually contains only one subject, the model struggles to combine multiple concepts in the same scene.
This lack of diversity prevents the model from learning distinguishing features between concepts. 
Attribute leakage happens when different subjects share attributes, causing the model to mix their identities. This is more common with class-similar subjects (e.g., two types of dogs) making it harder to generate their unique traits.
This problem degrades the fidelity of the generated images, particularly when fine-grained details across multiple concepts need to be preserved.

To tackle these challenges, we introduce Fuse-and-Refine (FaR), a personalized image generation method with two main contributions: Concept Fusion technique and Localized Refinement loss function. 
In Concept Fusion, we augment the reference set by separating reference subjects and recombining them in random positions to enhance diversity.
By enriching the training data with more varied compositions, this technique reduces overfitting and enhances learning of both single and multiple concepts simultaneously. 
To mitigate attribute leakage, we introduce Localized Refinement loss function. 
Our method preserves subject attributes by applying spatial segmentation constraints, ensuring that the attention map of each concept aligns with the correct region.
Both Concept Fusion and Localized Refinement are integrated into our training pipeline. By carefully fine-tuning specific modules, the model can learn new concepts effectively without losing previous knowledge. As a result, FaR improves multi-subject compositions and photorealism without incurring additional computational cost at the inference phase.

\section{Related Works}
\subsection{Text-to-image Diffusion Models}
Diffusion models \cite{diffusion, stablediffusion, fastdiffuser} have proven to be highly effective in learning data distributions, demonstrating impressive results in image synthesis and leading to various applications. Our primary experiments were conducted using Stable Diffusion \cite{stablediffusion}, a widely-used implementation of latent diffusion models (LDMs). StableDiffusion operates within the latent space of a pre-trained autoencoder, which reduces the dimensionality of data samples. This allows the diffusion model to exploit the compacted semantic features and visual patterns learned by the encoder. Several diffusion models \cite{layoutdiffusion, gligen, instancediffusion} offer layout guidance to give users fine-grained control over text-to-image generation. This enables the specification of subject placements, spatial arrangements, or compositional structures-features particularly beneficial for design prototyping, storytelling, or artistic creation where precise positioning is crucial. Despite these advances, diffusion models are often trained on extensive, general-purpose datasets, making it challenging to incorporate personalized or domain-specific concepts in the generated images. While layout-guided diffusion models \cite{layoutdiffusion, gligen, densediffusion, instancediffusion} provide strong control, they still fall short in referencing user-specific concepts.

In this work, we introduce an efficient approach for text-to-image personalization without using additional conditions, addressing the limitations of existing methods that struggle with incorporating specific, user-defined concepts. Our proposed method aims to maintain the generalization capability of the diffusion model while enabling precise personalization for individual needs.

\subsection{Text-to-image Personalization}
Stable Diffusion \cite{stablediffusion} based models have achieved remarkable progress, their capacity to adapt and faithfully represent unique, user-specific concepts remains constrained. Various techniques have emerged to address this issue. For instance, Textual Inversion \cite{textualinversion} optimizes specific embeddings, which are compact vector representations of text, to
associate them with a new visual concept (e.g., a new object, art style or person). 
Similarly, LoRA \cite{lora} avoids modifying the base model weights by inserting and training low-rank matrices in certain layers to reduce the number of training parameters \cite{mixofshow,loracomposer,blockwise-lora}. Both Textual Inversion and LoRA-based methods limit modifications to the base model weights to preserve prior knowledge. As a result, they may struggle to capture fine-grained details or distinguishing features of new concepts. 
Recent works \cite{customdiffusion, dreambooth, svdiff,subject-diffusion} refine the base model using a small set of exemplars, enabling it to learn custom subject details in diverse contexts.
These methods still face challenges in balancing underfitting, which reduces accuracy, and overfitting, which restricts diversity, due to the large amount number of parameters and limited training data.

To address these challenges, we introduce a new data augmentation strategy designed to mitigate overfitting. Furthermore, instead of fine-tuning all model weights, we systematically select key components to enhance adaptability. This approach allows the model to capture distinguishing features while maintaining prior knowledge.

\subsection{Multiple Concepts Generation}

Despite progress in diffusion models, ensuring text-to-image consistency across multiple concepts remains challenging. Various methods address this through spatial constraints, such as ControlNet based models \cite{controlnet,layoutdiffusion, gligen, densediffusion, instancediffusion} which utilize sketches, masks, or edges alongside text prompts to direct high-level features. However, diffusion models often struggle with complex relationships among multiple concepts, partly due to the limited representational capacity of the text encoder \cite{long-clip, structure-diffusion}. While some works \cite{attend-and-excite,divide-and-bind} adjust the latent space or cross-attention maps to refine compositional abilities, others \cite{structure-diffusion, SG-Adapter,linguistic-diffusion} focus on mitigating linguistic ambiguities.
Recent research in multi-concept generation has focused on personalizing concepts by learning each subject individually and combining them during inference. Some methods \cite{cones,conesv2, customdiffusion} fine-tune specific model components, while others improve the training process \cite{break-a-scene, concept-weaver} or propose data augmentation techniques \cite{svdiff,break-a-scene,mudi}. In contrast to previous works, our approach augments the training set by separating subjects from their backgrounds and recombining them into composite images. This strategy reduces overfitting and effectively supports multi-concept generation.

Some works \cite{mixofshow,loracomposer, concept-weaver} use spatial conditioning to guide the model in generating content for multiple subjects. This helps maintain the spatial relationships between the subjects and reduces the risk of missing any of them. Although these advancements are significant, existing methods still struggle attribute leakage when combining class-similar subjects. Unlike previous work, we incorporate Localized Refinement loss, which enforces spatial segmentation constraints and ensures that each concept’s attention map aligns with its designated region. As a result, our method significantly improves the composition of multiple subjects without requiring additional conditions, such as sketches or masks, during inference stage.


\section{Preliminaries}
\subsection{Text-to-image Diffusion Models}
Diffusion models gradually corrupt data with noise over multiple time steps and then learn to reverse this process to recover the original data distribution. Text-to-image diffusion models extend this concept by generating images from text descriptions within a compressed latent space. Text-to-image diffusion models aim to generate images from text descriptions by operating in a compressed latent space. Specifically, given training dataset $D$ consists of paired samples $(x,p)$ where $x$ represents image data and $p$ corresponds to its associated text description. A Variational Autoencoder (VAE) \(\mathcal{E}\) encodes an input image \(x\) into a latent representation \(z\). A text encoder then processes a text prompt \(p\) to produce a text embedding \(\tau(p)\). A neural network predicts the noise \(\epsilon\) added to the latent representation \(z_t\) at each diffusion step \(t\). The denoising network \(\epsilon_\theta(\cdot) \) is trained by minimizing the mean squared error between the predicted noise \(\epsilon_{\theta}( z_t, t, \tau(p))\) and the actual noise \(\epsilon\) sampled from a standard normal distribution:

\begin{equation}
    \mathcal{L}_{DM}(\theta;D) = \mathbb{E}_{z, p, t, \epsilon} \left[ \left\| \epsilon - \epsilon_{\theta} \left( z_t, t, \tau(p) \right) \right\|_2^2 \right]
    \label{LDMLoss}
\end{equation}

\subsection{Text-Conditioning via Cross-Attention Mechanism} \label{cross-attention}
\begin{figure}[t]
    \centering
    \includegraphics[width=\linewidth]{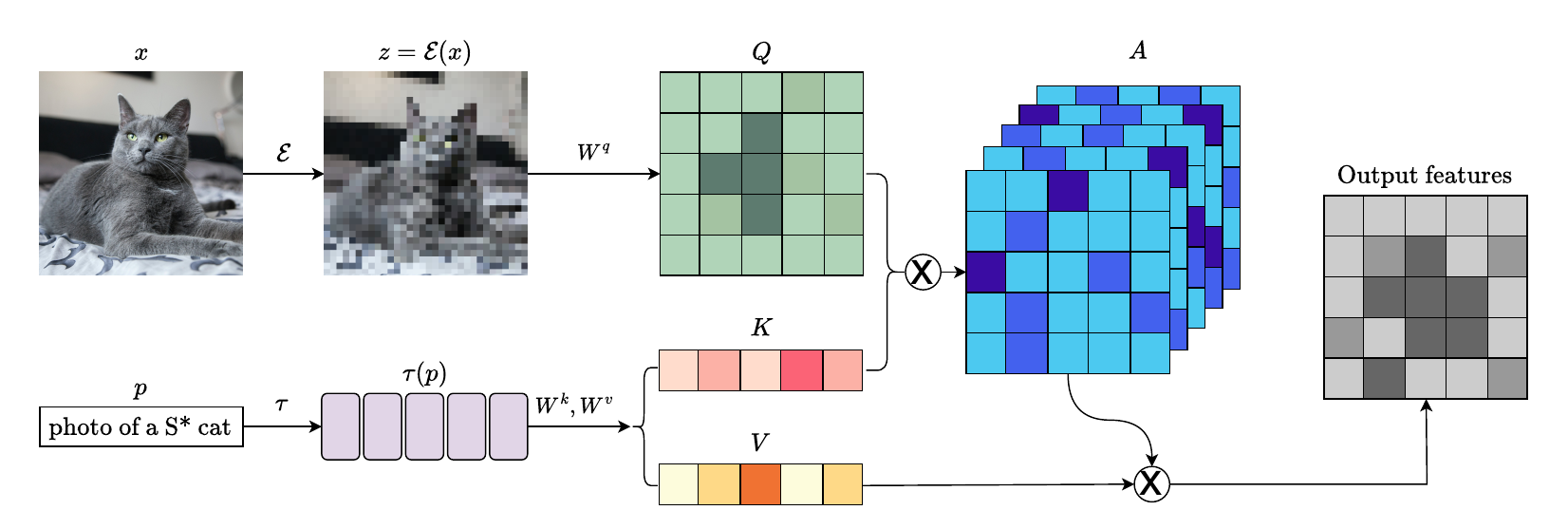}
    \caption{Illustration of Single-head Cross-Attention in Stable Diffusion. The image $x$ is processed through encoder \( \mathcal{E} \), generating a latent representation $z$. A text prompt $p$ is encoded into a text embedding \(\tau(p)\). $W^q$, $W^k$, and $W^v$ map the inputs to a query $Q$, key $K$, and value $V$ feature, respectively. 
    The cross-attention map $A$ is multiplied by $V$ to generate features that capture the interaction between image and text.}
    \label{fig_attention}
\end{figure}
The cross-attention mechanism in models like Stable Diffusion is essential for relating images to text conditions, enabling the Text-to-Image model to generate images that align consistently with the text descriptions. As depicted in Figure \ref{fig_attention}, we have a latent representation $z$ and a text embedding \( \tau(p) \), which are then input into the cross-attention layer. Following this, they are projected into Query ($Q$), Key ($K$), and Value ($V$) features by $W^q$, $W^k$ and $W^v$ matrix in the cross-attention block. Specifically, $Q$ is derived from the latent features of the noisy image, while $K$ and $V$ are projected from the text embedding. The cross-attention layer then computes the attention scores:

\begin{equation}
    A = \text{Softmax}\left(\frac{Q K^T}{\sqrt{d}}\right)
\end{equation}
where $d$ denotes the output dimension of the Query $Q$ and Key $K$ features. The output features $A \times V$ is a fused feature representation of both text and image, capturing the alignment between the two modalities. Each cell in the cross-attention map indicates how much a specific text token contributes to a spatial feature of the image, effectively distributing the textual information across the 2D latent code space. This allows the diffusion model to distribute and align the semantic content of the prompt with corresponding regions in the image, where $A[i,j,k]$ quantifies the flow of information from the $k$-th text token to the ($i,j$)-th latent pixel.

\section{Method}
\subsection{Concept Fusion for Multi-Subject Generation} \label{concept_fusion}

Empirical analysis shows that training concepts separately produces a model that performs well on individual concepts but struggles to generate images that combine multiple concepts effectively. Furthermore, with only a few training samples per concept (typically 3–5 images), the model is prone to overfitting. This overfitting often leads to language drift, where the fine-tuned model misaligns language inputs with generated images. Additionally, the outputs lack diversity in poses, shapes, and viewpoints, further limiting the model’s flexibility.

To address these problems, we present Concept Fusion, a data augmentation technique that enhances training diversity by \textit{incorporating multiple concepts into a single training sample}. In addition, we use the Stable Diffusion model \cite{stablediffusion} to generate \textit{prior images} that belong to the same class as the \textit{reference images}. First, we use the fine-grained class name of the reference subject (e.g., “border collie” or “chow chow”) to generate prior images. These images provide the model with prior knowledge about the subject’s general characteristics, helping it better capture variations in poses, shapes, and viewpoints. By leveraging both reference and prior images, we expand the training set, further enhancing its diversity in both generic and specific subject details.


\begin{figure}
    \centering
    \includegraphics[width=0.8\linewidth]{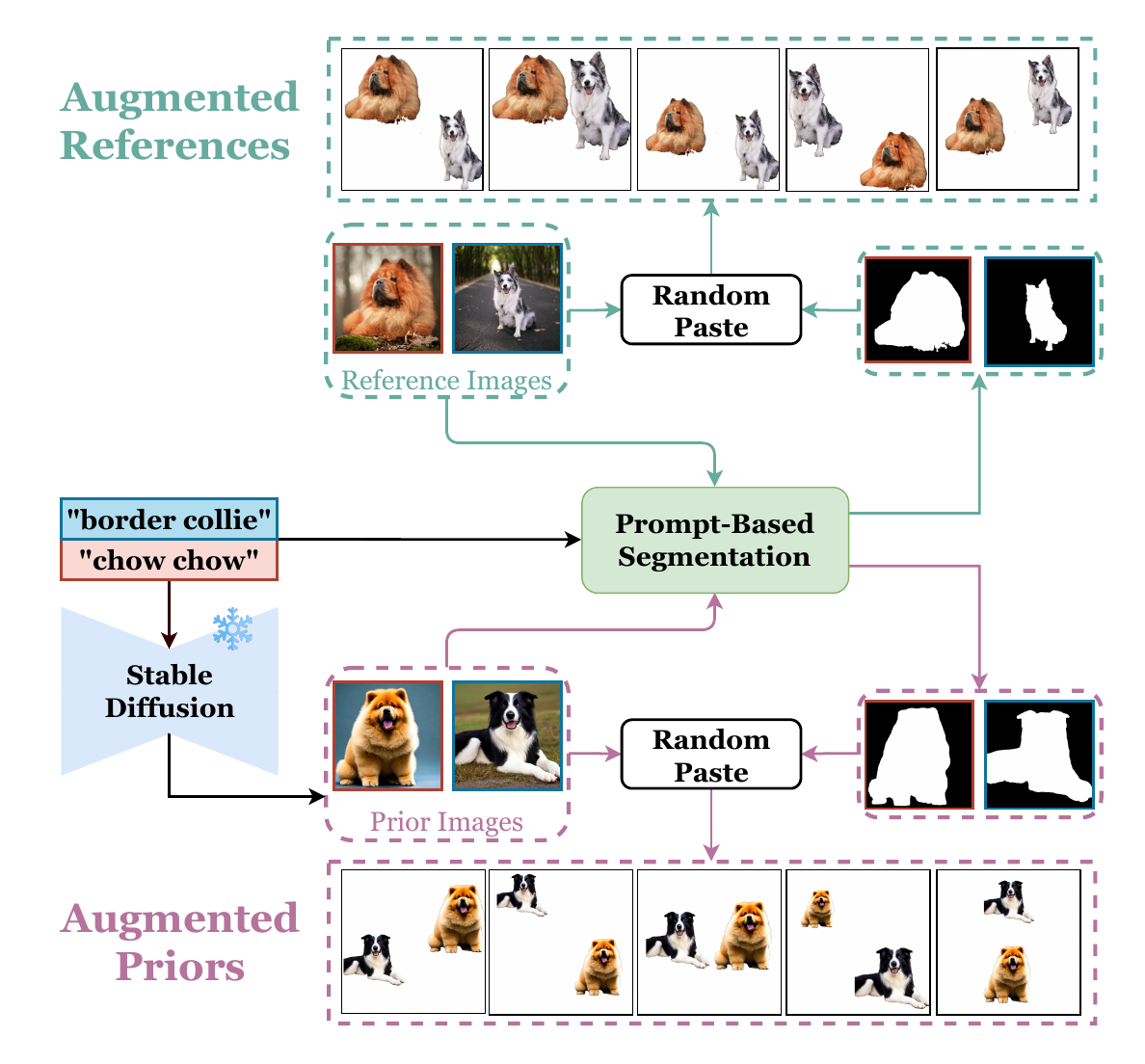}
    \caption{Overview of Concept Fusion. By separating each subject from the background and randomly positioning it on new composite samples, the Concept Fusion augmentation technique enhances the model's ability to differentiate between identities.}
    \label{fig_data_aug}
\end{figure}

After acquiring the reference and prior images, we automatically extract segmentation maps for user-specified subjects using Grounded SAM \cite{groundedsam} given the input images and the subject related prompts.
We then use these maps along with the images for data augmentation during training. Specifically, we create augmented images by randomly translating and resizing segmented subjects onto a simple background, allowing for occasional overlap between subjects. This transformation technique applies to both reference and prior images, producing \textit{augmented references} and \textit{augmented priors}. The combined set of reference images and augmented references is called $D_{ref}$, while the combined set of prior images and augmented priors is called $D_{prior}$.  The full workflow of Concept Fusion is illustrated in Figure \ref{fig_data_aug}.

\subsection{Training Pipeline}
\begin{figure*}
    \includegraphics[width=\linewidth]{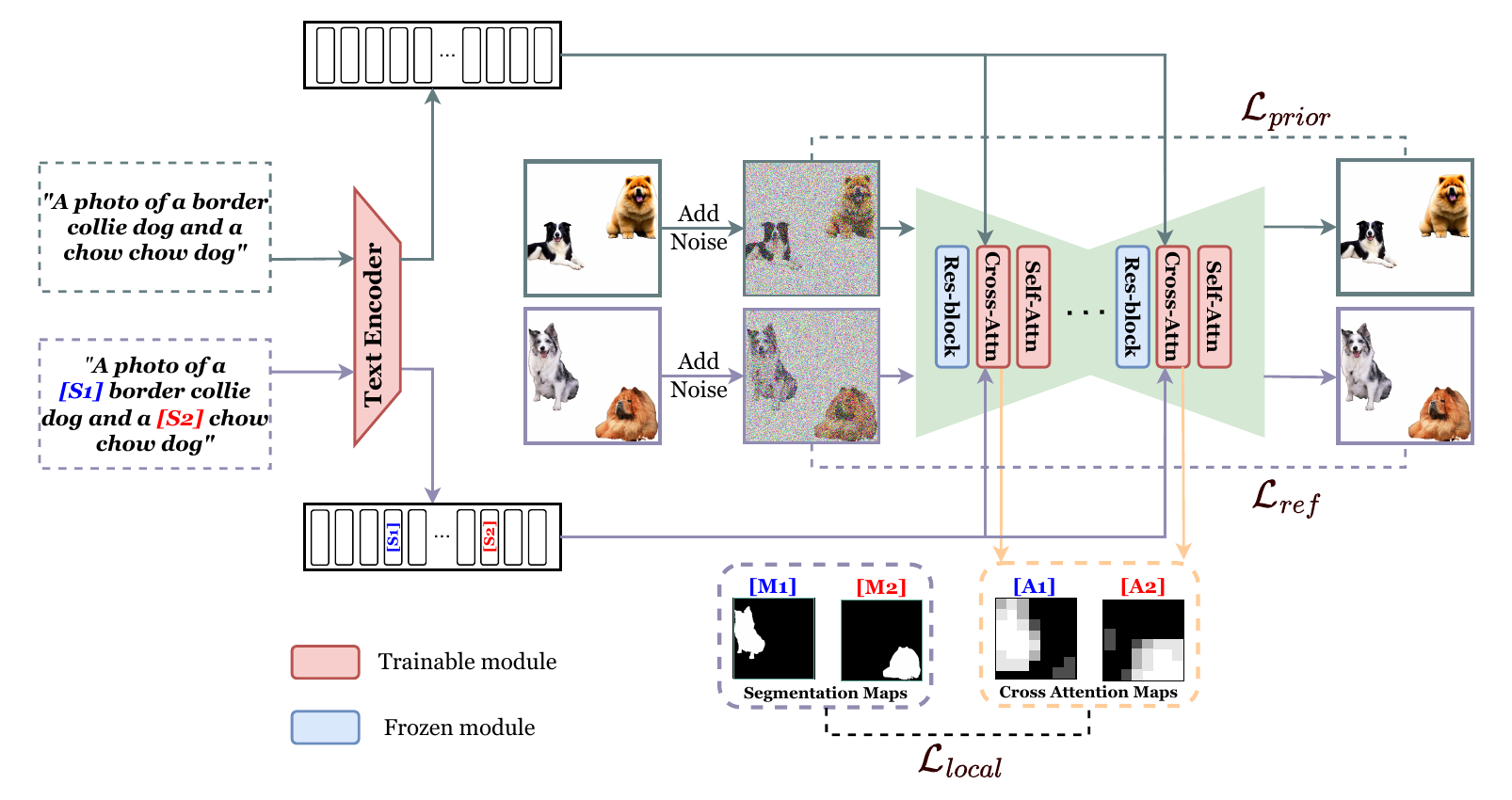}
    \caption{Our training pipeline is demonstrated using a subset of $k = 2$ subjects. For simplicity, we set the subject IDs as $C_1 = 1$ and $C_2 = 2$. During training, we simultaneously optimize the text encoder, self-attention layers, and cross-attention layers. This approach enables the model to learn detailed features of the new concepts while minimizing the loss of knowledge from the original model.}
    \label{fig_pipeline}
\end{figure*}


Our fine-tuning approach focuses on three key components: the cross-attention layers, the self-attention layers of the denoising network, and the text encoder. Fine-tuning the cross-attention layers improves the alignment between textual prompts and the generated visual features.
At the same time, adjusting the self-attention layers enhances the model's ability to capture complex spatial relationships and fine details that define the new concept, ensuring it focuses on relevant features during the denoising process. Additionally, refining the text encoder enables a more accurate representation of the semantic space for the new concept, ensuring better consistency with related classes. Collectively, these adjustments significantly enhance the model's ability to generate outputs that are both visually coherent and conceptually accurate.

Let $C=\{1,2,\dots,c\}$ be a set of new concept ids. At each training step, we randomly select a subset $\{C_1, C_2,\dots, C_k\} \subset C, k > 0$. Along with these subjects and training samples, we have corresponding masks $\{M_{C_{1}}, M_{C_{2}}, \dots, M_{C_{k}}\}$. Inspired by Textual Inversion \cite{textualinversion}, we define placeholder strings $S_{C_{1}},S_{C_{2}}, \dots, S_{C_{k}}$ to represent new concepts. We initialize concept embeddings of these placeholder strings with embeddings of their class names (e.g., “cat”, “person”), and undergo optimization to learn the new concept embeddings. Some works \cite{mudi, instructbooth} have shown that incorporating detailed descriptions before class names enhances the model’s ability to capture visual characteristics of subjects. In this paper, we adopt the prompt format “A photo of a $[S_{C_{1}}][\texttt{attributes}][\texttt{class name}]$ and ... $[S_{C_{k}}][\texttt{attributes}][\texttt{class name}]$” to guide the model distinguish similar subjects in the same training sample. For instance, we use a prompt “[$S_1$] \texttt{border collie} \texttt{dog}” instead of “[$S_1$] \texttt{dog}” or “[$S_2$] \texttt{pink backpack}” instead of “[$S_2$] \texttt{backpack}”. In the case of training with prior images, the prompt format does not include placeholder strings. The overall fine-tuning strategy is illustrated in Figure~\ref{fig_pipeline}.

\subsection{Localized Refinement Loss} \label{loss}
Personalizing the diffusion model to integrate multiple subjects remains challenging due to attribute leakage, particularly when working with subjects from the same class (e.g., two dogs). This occurs because the cross-attention maps tend to focus on all subjects at once \cite{customdiffusion, break-a-scene}. As discussed in Section \ref{cross-attention}, a cross-attention map allows the diffusion model to align and distribute the semantic content of a prompt with the corresponding regions in an image.  Here, \( A[i,j,k] \) quantifies the flow of information from the \( k \)-th text token to the \( (i,j) \)-th latent pixel. Ideally, the attention map for a subject token should concentrate solely on that subject's region, thereby preventing attribute leakage among different subjects.

To achieve this goal, our proposed Localized Refinement loss ensures that the model distinctly focuses on separate subject regions and effectively discourages overlapping attention maps between different subjects. We define the loss function as follows:
\begin{multline}
    \mathcal{L}_{\mathrm{sep}}(\theta;D_{\mathrm{ref}}) = \mathbb{E}_{C_i} \biggl[ \frac{1}{N^2} \sum_{h=1}^N \sum_{w=1}^N \Bigl[ M_{C_{i}} \odot \log A_{C_{i}}
    + (1-M_{C_{i}}) \odot \log(1-A_{C_{i}}) \Bigr]_{h,w} \biggr]
    \label{SL}
\end{multline}
where $A_{C_{i}}$ denotes the cross-attention maps corresponding to the text embedding of the concept $C_i$, $N$ is the size of attention matrix. We use cross-attention maps of size $16 \times 16$ at both the up and down cross-attention layers. This resolution has been empirically shown \cite{prompt-to-prompt, attend-and-excite} to effectively capture rich semantic information, offering a balance between computational efficiency and the retention of fine-grained details. By applying this spatial constraint, the model prevents attribute leakage and preserves high-fidelity details for each personalized concept.
The final training loss function of FaR is a combination of Equation \ref{LDMLoss} and Equation \ref{SL}:
\begin{equation}
        \mathcal{L}_{total}(\theta) = \underbrace{\mathcal{L}_{DM}(\theta;D_{ref})}_{\mathcal{L}_{ref}} + \mu \underbrace{\mathcal{L}_{DM}(\theta;D_{prior})}_{\mathcal{L}_{prior}} + \gamma \mathcal{L}_{local}(\theta;D_{ref})
\end{equation}
where $\mu$ and $\gamma$ are pre-defined scaling factors.

\section{Experiments}

\subsection{Experimental Setup}
\begin{figure}
    \centering
    \includegraphics[width=0.85\linewidth]{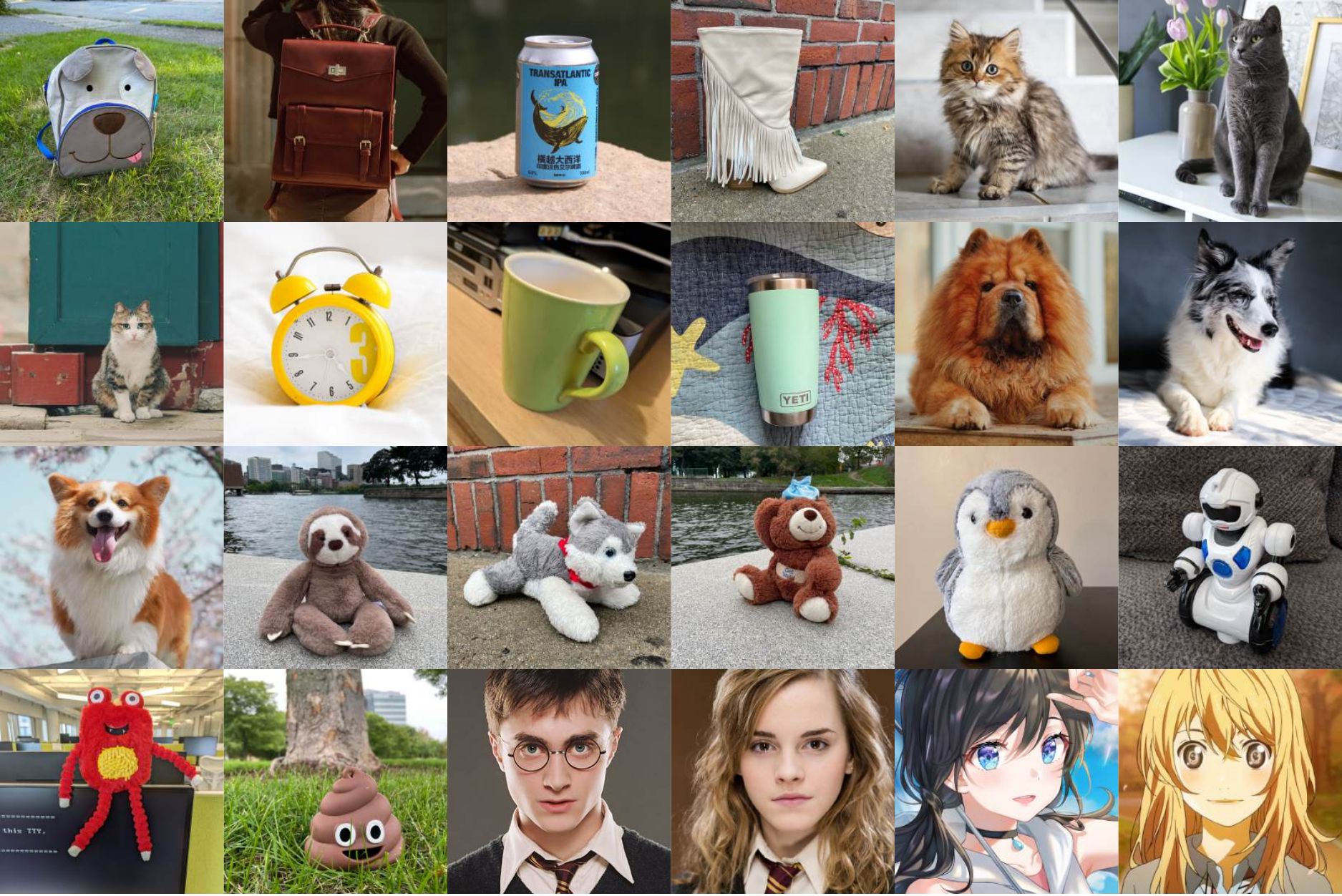}
    \caption{
    Our dataset of 24 subjects across humans, animals, and objects was used to evaluate personalization methods.}
    \label{fig_dataset}
\end{figure}

\textbf{Dataset.} We evaluate personalization methods using a dataset collected from three sources: the DreamBench dataset \cite{dreambooth}, the CustomConcept101 dataset \cite{customdiffusion}, and the Mix-of-show dataset \cite{mixofshow}. Our dataset includes 24 distinct concepts across various categories, such as humans, animals and objects, as shown in Figure \ref{fig_dataset}.

\textbf{Implementation details.}
All our experiments leverage pretrained Stable Diffusion V2.1 as the starting point for fine-tuning. We primarily focus on evaluating the ability of our method and other approaches to personalize two subjects. Specifically, in our method we use a learning rate of 2e-6 for 5000 steps. Concept Fusion data augmentation is applied throughout the training process with a rate of 0.5. The AdamW optimizer is employed with hyperparameters $\beta_1=0.9$, $\beta_2=0.99$ and weight decay set to 1e-2. We set the scaling factors for the overall loss function as $\mu=1.0$, $\gamma=0.04$.

We evaluate 1200 generated images across 16 combinations, using 5 evaluation prompts generated from ChatGPT for each combination. Each combination comprises two single-subject cases and one case featuring a pair of subjects. For every evaluation prompt, we generate 5 images. All methods are assessed using a fixed random seed of 42 throughout both the training and inference processes.

\textbf{Baselines.} We compare our method with several existing approaches, including Custom Diffusion \cite{customdiffusion}, Textual Inversion \cite{textualinversion}, ConesV2 \cite{conesv2}, Mix-of-Show \cite{mixofshow}, FreeCustom \cite{freecustom}, and Concept Conductor \cite{concept-conductor}. For a fair comparison, we use the official code implementation for each method and follow the recommended experimental settings provided by their authors.

\subsection{Main Result}

\begin{table*}
\centering
\caption{Quantitative comparisons.}
\label{table_results}
\begin{adjustbox}{width=\linewidth,center}
\begin{tabular}{|l|ccccc|ccccc|}
\hline
\multicolumn{1}{|c|}{} & \multicolumn{5}{c|}{Single-Subject Fidelity}                                                                                                                                                                           & \multicolumn{5}{c|}{Multi-Subject Fidelity}                                                                                                                                                                            \\ \cline{2-11} 
\multicolumn{1}{|c|}{Method} & \multicolumn{2}{c}{\begin{tabular}[c]{@{}c@{}}Image\\ Alignment\end{tabular}} & \multicolumn{2}{c}{\begin{tabular}[c]{@{}c@{}}Text\\ Alignment\end{tabular}} & \begin{tabular}[c]{@{}c@{}}Image\\ Quality\end{tabular} & \multicolumn{2}{c}{\begin{tabular}[c]{@{}c@{}}Image\\ Alignment\end{tabular}} & \multicolumn{2}{c}{\begin{tabular}[c]{@{}c@{}}Text\\ Alignment\end{tabular}} & \begin{tabular}[c]{@{}c@{}}Image\\ Quality\end{tabular} \\
\multicolumn{1}{|c|}{}                & D\&C-DS↑                               & D\&C-DINO↑                            & \begin{tabular}[c]{@{}c@{}}IR↑\end{tabular}   & CLIP↑             & CLIP-IQA↑                                                & D\&C-DS↑                              & D\&C-DINO↑                            & \begin{tabular}[c]{@{}c@{}}IR↑\end{tabular}   & CLIP↑             & CLIP-IQA↑                                                \\ \hline
Textual Inversion \cite{textualinversion}     & 0.706                                 & 0.725                                 & -1.293                                                   & 21.291            & 0.835                                                   & 0.074                                 & 0.097                                 & -1.743                                                   & 18.241            & 0.814                                                   \\
Custom Diffusion \cite{customdiffusion}       & 0.765                                 & 0.780                                 & {\ul 1.025}                                              & {\ul 26.178}      & 0.904                                                   & 0.335                                 & 0.386                                 & {\ul 0.707}                                              & 26.293            & 0.913                                                   \\
FreeCustom \cite{freecustom}             & 0.671                                 & 0.687                                 & -0.186                                                   & 23.210            & 0.819                                                   & 0.259                                 & 0.286                                 & -0.406                                                   & 22.214            & 0.794                                                   \\
Mix-of-Show \cite{mixofshow}           & {\ul 0.791}                           & {\ul 0.786}                           & 0.470                                                    & 24.967            & 0.913                                                   & {\ul 0.480}                           & {\ul 0.471}                           & 0.563                                                    & \textbf{27.282}   & 0.890                                                   \\
Concept Conductor \cite{concept-conductor}     & 0.542                                 & 0.570                                 & 0.595                                                    & 26.023            & {\ul 0.917}                                                   & 0.449                                 & 0.456                                 & 0.685                                                    & {\ul 26.869}      & 0.914                                                   \\
Cones-V2 \cite{conesv2}              & 0.624                                 & 0.678                                 & 0.792                                                    & \textbf{26.283}   & \textbf{0.921}                                          & 0.233                                 & 0.257                                 & 0.198                                                    & 25.170            & \textbf{0.938}                                          \\ \hline
FaR (Ours)                   & \textbf{0.849}                        & \textbf{0.847}                        & \textbf{1.062}                                           & 25.604            & 0.910                                             & \textbf{0.664}                        & \textbf{0.627}                        & \textbf{0.934}                                           & 25.607            & {\ul 0.915}                                             \\ \hline
\end{tabular}
\end{adjustbox}
\end{table*}

\textbf{Quantitative evaluation.} We evaluate the performance of personalization methods quantitatively using metrics for both single-subject and multi-subject fidelity. The evaluation focuses three key aspects: image alignment, text alignment, and image quality. For image alignment, we utilize D\&C scores \cite{mudi} to assess the preservation of visual details for each subject and the accuracy in generating the correct number of subjects. Specifically, D\&C-DS employs the DreamSim model \cite{dreamsim}, while D\&C-DINO leverages the DINOv2 model \cite{dinov2} to extract image embeddings, which are then used to compute similarity scores. To evaluate text alignment, we utilize CLIP score \cite{clip} and ImageReward (IR) \cite{imagereward} to assess how effectively the generated images match the prompts. Additionally, we use CLIP-IQA \cite{clip-iqa} to evaluate the overall quality of the generated images. 

The results, summarized in Table \ref{table_results}, show that FaR significantly outperform other sate-of-the-art methods across all key metrics such as D\&C-DS, D\&C-DINO, and IR. Although our method results in a lower CLIP score, this is because the CLIP model primarily focuses on global semantics and does not explicitly capture fine-grained details accurately. For the CLIP-IQA metric, our model achieves a score of 0.91, which is slightly lower than Cones-V2 (0.921) for single-subject generation, with a similar trend observed for multi-subject cases. Despite this minor difference, a score above 0.91 indicates that our generated images maintain high quality and are suitable for real-world applications.

\begin{figure}
    \centering
    \includegraphics[width=\linewidth]{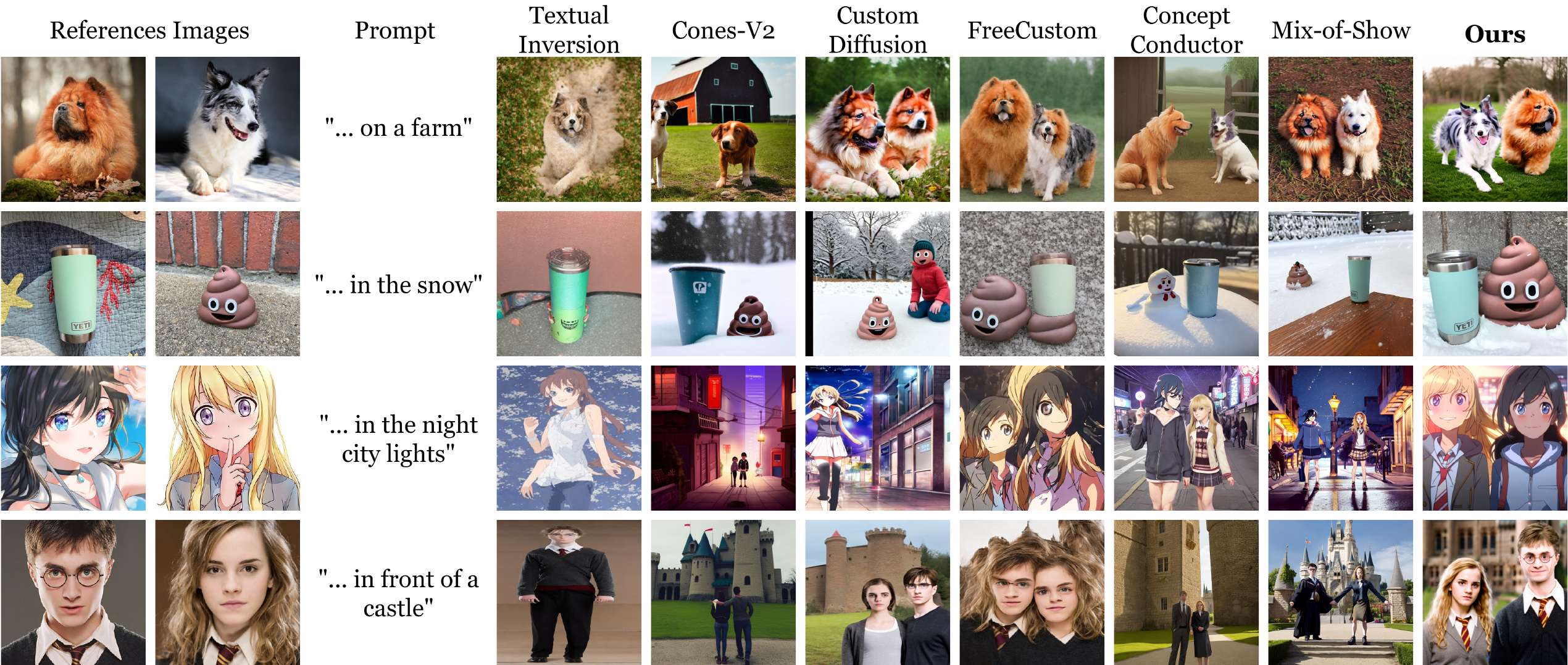}
    \caption{\textbf{Qualitative Comparison of Multi-Concept Generation.} Our method outperforms others by consistently preserving subject identities, spatial relationships, and accurately adapting subjects to new scenes in multi-subject scenarios.}
    \label{fig_result_multi}
\end{figure}

\textbf{Qualitative comparison.}
The results, as shown in Figure \ref{fig_result_multi}, demonstrate the stability of our method in generating various concept combinations, even for class-similar subjects, such as two dogs. Unlike existing methods that may struggle to distinguish and accurately compose similar subjects, our method consistently maintains clear subject separation and preserves their integrity. As shown in the figure, our generated images not only achieve high quality but also effectively prevent attribute leakage between subjects.

\begin{figure}
    \centering
    \includegraphics[width=\linewidth]{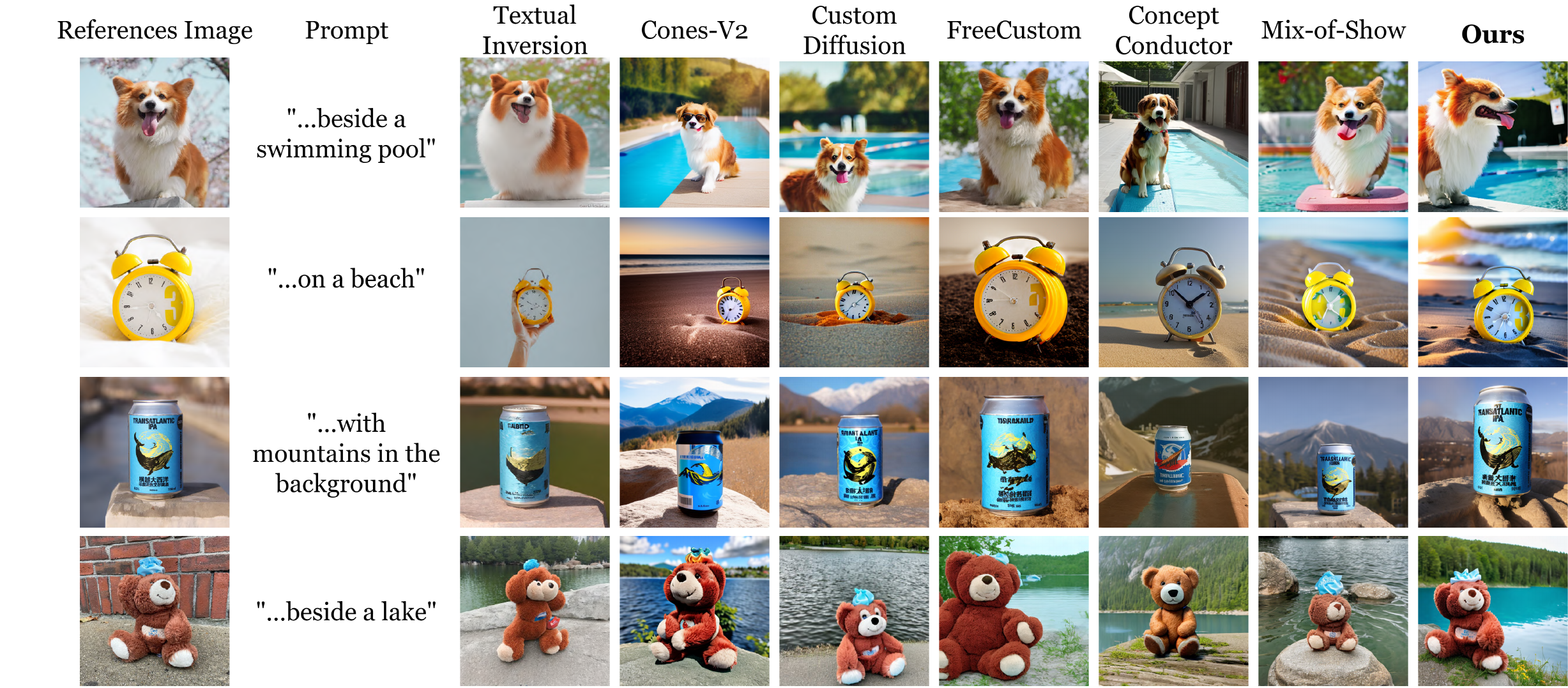}
    \caption{\textbf{Qualitative Comparison of Single-Concept Generation.} 
    Our approach (last column) outperforms others by generating visually consistent, contextually accurate representations while preserving target context and reference appearance.}
    \label{fig_result_single}
\end{figure}

In addition to strong performance in multi-concept scenarios, our method also proves to be highly effective in single-concept generation tasks. As illustrated in Figure \ref{fig_result_single}, our approach ensures stability and fidelity, allowing the generated subjects to retain fine-grained details and semantic consistency. Overall, these results highlight the versatility of our method, making it not only robust for complex multi-concept scenarios but also highly reliable for generating high-quality outputs in single-concept task.

\subsection{Ablation Studies}
\begin{figure}
    \centering
    \includegraphics[width=0.8\linewidth]{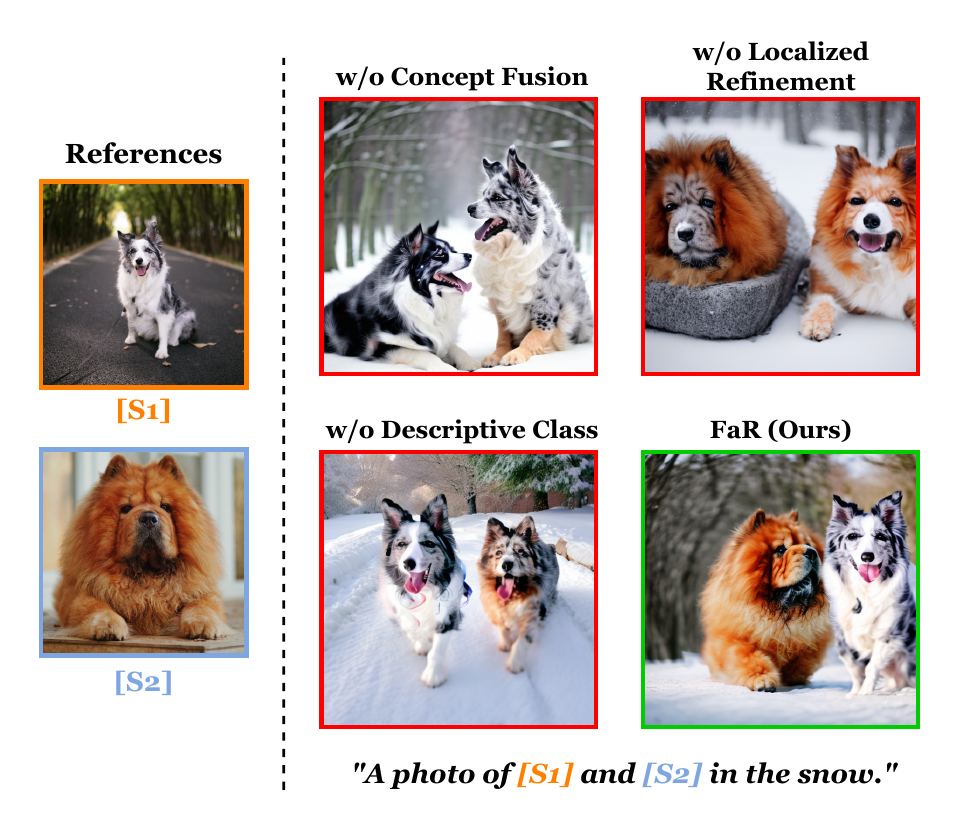}
    \caption{
    Ablation results show our full pipeline excels in fidelity and coherence, effectively combining [S1] and [S2] with accurately detailed features.}
    \label{fig_ablation}
\end{figure}

\textbf{Without Concept Fusion. }
When Concept Fusion is omitted, the model faces significant challenges in generating images that integrate multiple subjects while preserving their individual visual characteristics as shown in Figure \ref{fig_ablation}. This issue arises because training subjects in isolation causes the model to over-specialize on each subject, making it difficult to disentangle their distinct features when combined. Table \ref{table_ablation} shows that the model struggles to generalize, frequently generating images where subjects lose their identity.

\begin{table*}
\centering
\caption{Results on ablation studies.}
\begin{adjustbox}{width=\linewidth,center}
\begin{tabular}{|l|ccccc|ccccc|}
\hline
\multicolumn{1}{|c|}{} & \multicolumn{5}{c|}{Single-Subject Fidelity}                                                                                                                                                                           & \multicolumn{5}{c|}{Multi-Subject Fidelity}                                                                                                                                                                            \\ \cline{2-11} 
\multicolumn{1}{|c|}{Method} & \multicolumn{2}{c}{\begin{tabular}[c]{@{}c@{}}Image\\ Alignment\end{tabular}} & \multicolumn{2}{c}{\begin{tabular}[c]{@{}c@{}}Text\\ Alignment\end{tabular}} & \begin{tabular}[c]{@{}c@{}}Image\\ Quality\end{tabular} & \multicolumn{2}{c}{\begin{tabular}[c]{@{}c@{}}Image\\ Alignment\end{tabular}} & \multicolumn{2}{c}{\begin{tabular}[c]{@{}c@{}}Text\\ Alignment\end{tabular}} & \begin{tabular}[c]{@{}c@{}}Image\\ Quality\end{tabular} \\
\multicolumn{1}{|c|}{}                 & D\&C-DS↑                               & D\&C-DINO↑                             & \begin{tabular}[c]{@{}c@{}}IR↑\end{tabular}   & CLIP↑              & CLIP-IQA↑                                                & D\&C-DS↑                               & D\&C-DINO↑                             & \begin{tabular}[c]{@{}c@{}}IR↑\end{tabular}   & CLIP↑              & CLIP-IQA↑                                                \\ \hline
w/o Concept Fusion     & {\ul 0.820}                           & {\ul 0.833}                          & 0.217                                                    & {\ul 23.727}            & 0.888                                                   & 0.155                                 & 0.169                                 & -0.386                                                   & 22.042            & 0.896                                             \\
w/o Localized Refinement       & 0.745                                 & 0.763                                 & 0.105                                                    & 23.559           & 0.895                                                  & 0.467                                 &  0.470                                 & -0.409                                                   & 22.349            & 0.892
                    \\
w/o Descriptive Class       & 0.721                                 & 0.734                                 & {\ul 0.256}                                                    & 23.699           & {\ul 0.899}                                                  & {\ul 0.485}                                 & {\ul 0.502}                                 & -0.029                                                   & {\ul 23.068}            & {\ul 0.902}  
\\ \hline
FaR (Ours)             & \textbf{0.849}                        & \textbf{0.847}                        & \textbf{1.062}                                           & \textbf{25.604}   & \textbf{0.910}                                          & \textbf{0.664}                        & \textbf{0.627}                        & \textbf{0.934}                                           & \textbf{25.607}   & \textbf{0.915}                                          \\ \hline
\end{tabular}
\end{adjustbox}
\label{table_ablation}
\end{table*}

\textbf{Without Localized Refinement. }
To evaluate the impact of the proposed Localized Refinement, we conducted an ablation study by removing this component from the training process. Without Localized Refinement, the model could not effectively enforce separation between attention maps corresponding to different concepts. The absence of Localized Refinement caused identity blending between subjects in multi-concept scenarios. For example, as shown in Figure \ref{fig_ablation}, the generated images often exhibited overlapping regions where features of one subject blended into another. This blending not only diminished the visual clarity of the output but also affected the semantic alignment between the textual description and the image.

\textbf{Without Descriptive Class.} Figure \ref{fig_ablation} illustrates that employing descriptive classes to represent subjects enhances the preservation of their details. Additionally, Table \ref{table_ablation} further substantiates that this approach improves subject fidelity.

\section{CONCLUSION}
In this paper, we introduce Fuse-and-Refine (FaR), a novel fine-tuning approach designed to tackle critical challenges such as overfitting and attribute leakage in personalized text-to-image generation, particularly when dealing with multiple class-similar subjects. The extensive quantitative and qualitative evaluations demonstrate the effectiveness of FaR in generating high-fidelity images with multiple user-defined subjects. Our approach consistently outperform existing methods in terms of reducing identity mixing, maintaining subject clarity, and producing photorealistic results, even in complex multi-concept scenarios. In summary, our proposed method advances personalized text-to-image generation by tackling the core limitations of multi-concept composition, paving the way for more flexible and reliable image synthesis.

\bibliographystyle{splncs04}
\bibliography{ref}
\end{document}